\documentclass[letterpaper, 10 pt, conference]{class/ieeeconf}  

\IEEEoverridecommandlockouts                              
\usepackage{verbatim} 
\usepackage{graphicx}
\usepackage{amsmath}
\usepackage{amssymb}
\usepackage{latexsym}
\usepackage{url}
\usepackage{cite}
\usepackage{relsize}

\usepackage[table,dvipsnames]{xcolor}
\usepackage{colortbl}

\usepackage{calc}

\usepackage{multirow}
\usepackage{afterpage}
\usepackage{ifthen}
\usepackage{graphicx}
\usepackage{algpseudocode,algorithm,algorithmicx}
\usepackage{subfigure}
\usepackage{flushend}
\usepackage{epstopdf}
\usepackage{stackengine}
\usepackage{gensymb}
\usepackage[colorlinks=true, bookmarks=false, linkcolor=blue, urlcolor=blue, citecolor=blue]{hyperref} 
\usepackage{booktabs}
\usepackage{makecell}
\usepackage{hhline}
\usepackage{courier}
\usepackage{lipsum}
\usepackage{copyrightbox} 

\usepackage{caption} 

\hypersetup{pdfstartview={FitH}}
\usepackage{titlesec}
\renewcommand{\thesubsection}{\Alph{subsection}}
\titleformat{\subsection}
  {\normalfont\itshape} 
  {\thesubsection.}   
  {0pt}                
  {}                   

\title{\LARGE \bf SurgCUT3R: Surgical Scene-Aware Continuous Understanding of Temporal 3D Representation }

\author{Kaiyuan Xu$^{1}$, Fangzhou Hong$^{2}$, Daniel Elson$^{1}$ and Baoru Huang$^{\dagger}$$^{1,3}$
\thanks{$^{1}$The Hamlyn Centre for Robotic Surgery, Imperial College London, SW7 2AZ, UK.}%
\thanks{$^{2}$S-Lab, College of Computing and Data Science, Nanyang Technological University, Singapore 639798.}%
\thanks{$^{3}$Department of Computer Science, University of Liverpool, L69 7ZX, UK. {\tt\small Baoru.Huang@liverpool.ac.uk} }
\thanks{$^{\dagger}$ Corresponding author}}

\begin{document}

\newtheorem{problem}{Problem}
\newtheorem{lemma}{Lemma}
\newtheorem{theorem}[lemma]{Theorem}
\newtheorem{claim}{Claim}
\newtheorem{corollary}[lemma]{Corollary}
\newtheorem{definition}[lemma]{Definition}
\newtheorem{proposition}[lemma]{Proposition}
\newtheorem{remark}[lemma]{Remark}
\newenvironment{LabeledProof}[1]{\noindent{\it Proof of #1: }}{\qed}

\def\beq#1\eeq{\begin{equation}#1\end{equation}}
\def\bea#1\eea{\begin{align}#1\end{align}}
\def\beg#1\eeg{\begin{gather}#1\end{gather}}
\def\beqs#1\eeqs{\begin{equation*}#1\end{equation*}}
\def\beas#1\eeas{\begin{align*}#1\end{align*}}
\def\begs#1\eegs{\begin{gather*}#1\end{gather*}}

\newcommand{\poly}{\mathrm{poly}}
\newcommand{\eps}{\epsilon}
\newcommand{\e}{\epsilon}
\newcommand{\polylog}{\mathrm{polylog}}
\newcommand{\rob}[1]{\left( #1 \right)} 
\newcommand{\sqb}[1]{\left[ #1 \right]} 
\newcommand{\cub}[1]{\left\{ #1 \right\} } 
\newcommand{\rb}[1]{\left( #1 \right)} 
\newcommand{\abs}[1]{\left| #1 \right|} 
\newcommand{\zo}{\{0, 1\}}
\newcommand{\zonzo}{\zo^n \to \zo}
\newcommand{\zokzo}{\zo^k \to \zo}
\newcommand{\zot}{\{0,1,2\}}
\newcommand{\en}[1]{\marginpar{\textbf{#1}}}
\newcommand{\efn}[1]{\footnote{\textbf{#1}}}
\newcommand{\vecbm}[1]{\boldmath{#1}} 
\newcommand{\uvec}[1]{\hat{\vec{#1}}}
\newcommand{\thv}{\vecbm{\theta}}
\newcommand{\junk}[1]{}
\newcommand{\var}{\mathop{\mathrm{var}}}
\newcommand{\rank}{\mathop{\mathrm{rank}}}
\newcommand{\diag}{\mathop{\mathrm{diag}}}
\newcommand{\tr}{\mathop{\mathrm{tr}}}
\newcommand{\acos}{\mathop{\mathrm{acos}}}
\newcommand{\atantwo}{\mathop{\mathrm{atan2}}}
\newcommand{\SVD}{\mathop{\mathrm{SVD}}}
\newcommand{\quadf}{\mathop{\mathrm{q}}}
\newcommand{\linterp}{\mathop{\mathrm{l}}}
\newcommand{\sgn}{\mathop{\mathrm{sign}}}
\newcommand{\sym}{\mathop{\mathrm{sym}}}
\newcommand{\avg}{\mathop{\mathrm{avg}}}
\newcommand{\mean}{\mathop{\mathrm{mean}}}
\newcommand{\erf}{\mathop{\mathrm{erf}}}
\newcommand{\grad}{\nabla}
\newcommand{\R}{\mathbb{R}}
\newcommand{\defeq}{\triangleq}
\newcommand{\dims}[2]{[#1\!\times\!#2]}
\newcommand{\sdims}[2]{\mathsmaller{#1\!\times\!#2}}
\newcommand{\udims}[3]{#1}
\newcommand{\udimst}[4]{#1}
\newcommand{\com}[1]{\rhd\text{\emph{#1}}}
\newcommand{\ind}{\hspace{1em}}
\newcommand{\argmin}[1]{\underset{#1}{\operatorname{argmin}}}
\newcommand{\floor}[1]{\left\lfloor{#1}\right\rfloor}
\newcommand{\step}[1]{\vspace{0.5em}\noindent{#1}}
\newcommand{\quat}[1]{\ensuremath{\mathring{\mathbf{#1}}}}
\newcommand{\norm}[1]{\left\lVert#1\right\rVert}
\newcommand{\ignore}[1]{}
\newcommand{\specialcell}[2][c]{\begin{tabular}[#1]{@{}c@{}}#2\end{tabular}}
\newcommand*\Let[2]{\State #1 $\gets$ #2}
\newcommand{\algorithmicbreak}{\textbf{break}}
\newcommand{\Break}{\State \algorithmicbreak}
\newcommand{\ra}[1]{\renewcommand{\arraystretch}{#1}}

\renewcommand{\vec}[1]{\mathbf{#1}} 

\algdef{S}[FOR]{ForEach}[1]{\algorithmicforeach\ #1\ \algorithmicdo}
\algnewcommand\algorithmicforeach{\textbf{for each}}
\algrenewcommand\algorithmicrequire{\textbf{Require:}}
\algrenewcommand\algorithmicensure{\textbf{Ensure:}}
\algnewcommand\algorithmicinput{\textbf{Input:}}
\algnewcommand\INPUT{\item[\algorithmicinput]}
\algnewcommand\algorithmicoutput{\textbf{Output:}}
\algnewcommand\OUTPUT{\item[\algorithmicoutput]}
\maketitle
\thispagestyle{empty}
\pagestyle{empty}

\begin{abstract}
Reconstructing surgical scenes from monocular endoscopic video is critical for advancing robotic-assisted surgery. However, the application of state-of-the-art general-purpose reconstruction models is constrained by two key challenges: the lack of supervised training data and performance degradation over long video sequences. To overcome these limitations, we propose SurgCUT3R, a systematic framework that adapts unified 3D reconstruction models to the surgical domain. Our contributions are threefold. First, we develop a data generation pipeline that exploits public stereo surgical datasets to produce large-scale, metric-scale pseudo-ground-truth depth maps, effectively bridging the data gap. Second, we propose a hybrid supervision strategy that couples our pseudo-ground-truth with geometric self-correction to enhance robustness against inherent data imperfections. Third, we introduce a hierarchical inference framework that employs two specialized models to effectively mitigate accumulated pose drift over long surgical videos: one for global stability and one for local accuracy. Experiments on the SCARED and StereoMIS datasets demonstrate that our method achieves a competitive balance between accuracy and efficiency, delivering near state-of-the-art but substantially faster pose estimation and offering a practical and effective solution for robust reconstruction in surgical environments. Project page: \href{https://chumo-xu.github.io/SurgCUT3R-ICRA26/}{https://chumo-xu.github.io/SurgCUT3R-ICRA26/}. 
\end{abstract}


\section{INTRODUCTION}
\label{Sec:Intro}

Reconstruction of the surgical scene from monocular endoscopic video is a crucial task in advancing robotic-assisted surgery\cite{RAS, chen2025surgicalgs, xu2022self}. By creating a dense Reconstruction model of the observed tissues and instruments, it enables a range of downstream applications, including intraoperative navigation~\cite{navigation, role, huang2022simultaneous}, robotic surgery automation\cite{envisors, zhu2026robust}, and virtual reality simulation\cite{VR, huang2023detecting, vir}. This problem has been long studied at the intersection of computer vision and medical imaging, building upon foundational areas such as monocular depth estimation\cite{DAsurgery, monolot}, Structure-from-Motion (SfM)\cite{endosfm, endoDaM}, and Simultaneous Localization and Mapping (SLAM)~\cite{emdqslam, endoslam}.

Despite decades of progress, achieving robust, scale-consistent reconstruction from monocular endoscopic video remains an open challenge. Classical SfM or SLAM pipelines~\cite{endosfm,endoDaM, endoslam}, while mature in rigid and well-textured environments, often break down in surgical settings due to non-rigid tissue deformation, frequent occlusions by instruments, and texture-poor surfaces. Meanwhile, general-purpose learning-based methods for monocular depth estimation \cite{VDA, DAV2, DA} have made rapid strides, but they suffer from a severe domain gap when applied to surgical data. As a result, they typically fail to deliver the geometrically-consistent and scale-consistent reconstructions that meet the requirements of clinical applications.

\begin{figure}[t]
    \centering
    \includegraphics[width=0.98\columnwidth]{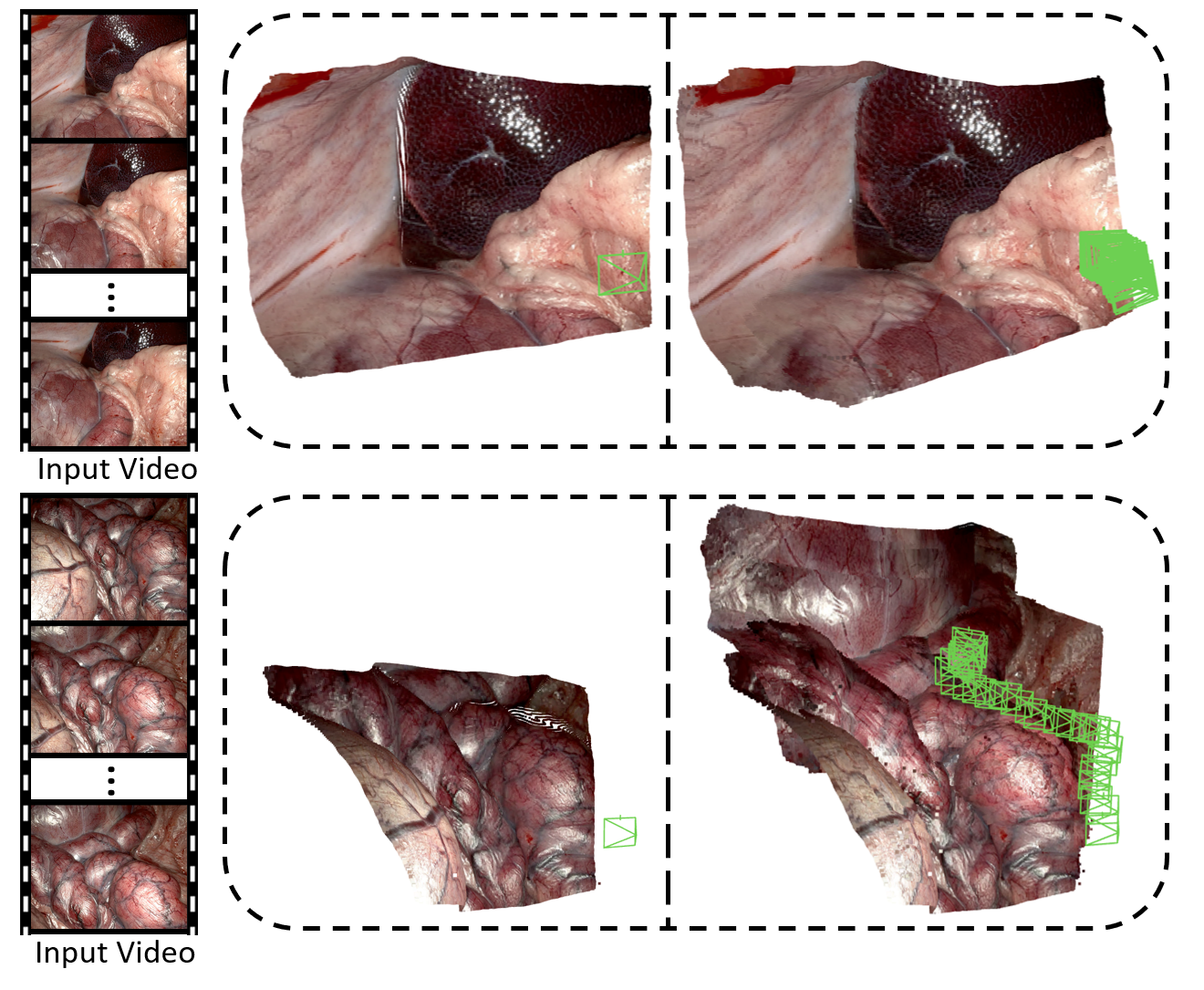}
    \vspace{1mm}
    \caption{\textbf{Qualitative results of 3D reconstruction.} With videos (small images) as input, this figure shows the reconstruction from the first frame (large images left) and the accumulated 3D model from multiple frames (large images right). This alignment between the single-frame and multi-frame reconstruction results highlights the geometric consistency of our method.}
    \label{fig:reconstruction_examples}
\end{figure}

In parallel, recent advances in large-scale learning-based models for general-purpose 3D reconstruction, such as those based-on DUSt3R\cite{dust3r}, have shown transformative potential. Unified frameworks like CUT3R \cite{cut3r} have demonstrated remarkable success in reconstructing diverse scenes from online monocular videos. However, adapting this state-of-the-art (SOTA) technology to the surgical domain is impeded by two fundamental barriers. First, the success of these models is based on vast datasets with high-quality, ground-truth (GT) 3D data for supervision, creating a significant data gap for the surgical domain where such data is difficult to acquire. Second, the autoregressive architecture of these models (e.g. CUT3R\cite{cut3r}), though effective for short clips, degrade over the long, continuous surgical video streams, leading to a scalability gap manifested as accumulated pose drift.

To address these challenges, we propose {SurgCUT3R}, a systematic framework that adapts an SOTA unified reconstruction model to the surgical domain. This work specifically addresses two fundamental challenges: (i) How can we effectively train a supervised model in the absence of real GT data? (ii) How can we architect a long-term inference pipeline that preserves the model's accuracy while mitigating pose drift?

Our main contributions are summarized as follows:
\begin{itemize}
    \item We develop a scalable data generation pipeline that leverages public stereo surgical videos to produce large-scale pseudo-GT depth maps with scale-consistency, bridging the data gap for supervised model training.
    \item We introduce a {hybrid supervision strategy} that combines direct supervision from our pseudo-GT with a comprehensive geometric self-correction mechanism, enhancing robustness against inherent data imperfections.
    \item We design a {hierarchical inference framework} that employs two specialized models to effectively mitigate pose drift, enabling stable camera tracking over long surgical video sequences.
\end{itemize}

\section{Related Work}
\label{sec:rw}

\subsection{Learning-Based Dense Correspondence for Reconstruction}
Recent advances in learning-based dense correspondence have been largely driven by transformer architectures, such as CroCo\cite{croco}, which perform the task of cross-view completion. This pre-training paradigm enabled the development of DUSt3R \cite{dust3r}, a seminal work that recasts the problem by directly regressing dense 3D pointmaps from an image pair. By predicting two pointmaps in one coordinate frame, DUSt3R\cite{dust3r} implicitly solves for relative pose and geometry but requires an offline, optimization-based global alignment for multi-view scenes. The DUSt3R\cite{dust3r} framework inspired several specialized extensions. MASt3R\cite{mast3r} improved matching precision by adding a head to predict dense features, improving performance on correspondence tasks like visual localization. Concurrently, MonST3R \cite{monst3r} adapted the model to handle dynamic scenes by fine-tuning it on video data with moving objects. To overcome the offline, pairwise nature of these models, Spann3R \cite{spann3r} introduced an online and incremental approach. It incorporates an external spatial memory to store past geometric information, allowing each new frame's pointmap to be regressed directly into a consistent global coordinate system without optimization. Building on this concept of continuous reconstruction, CUT3R \cite{cut3r} proposed a model with a persistent state that is not simply a cache of past observations. This learned state representation captures powerful scene priors, enabling the model to not only reconstruct observed regions online but also to infer and generate geometry for unobserved areas by querying the state with virtual camera views.

\subsection{SLAM-based Methods for Long-Sequence Consistency}
To mitigate the inherent drift of reconstruction models over long sequences, several works have integrated them into SLAM-based systems to ensure global consistency. MASt3R-SLAM~\cite{mast3rslam} pioneered this direction by building a real-time, keyframe-based SLAM system upon the MASt3R\cite{mast3r} prior, treating the two-view model as a robust front-end for a classic SLAM back-end with pose graph optimization. SLAM3R \cite{slam3r} pursued a fully neural approach, avoiding explicit camera parameters and using separate networks to generate local reconstructions and register them into a global model. To address the challenge of casual dynamic videos, MegaSaM \cite{megasam} extended a deep visual SLAM framework by integrating monocular depth priors and motion probability maps into a differentiable bundle adjustment layer and using excellent optimization methods to enhance robustness for low-parallax scenarios. 

\subsection{Reconstruction for Surgical Scenes}
The 3D reconstruction of surgical scenes has rapidly developed. Early self-supervised methods like AF-SfMLearner~\cite{afsfm} addressed specific challenges such as brightness inconsistency but relied on separate networks for depth and motion estimation. The advent of foundation models shifted the focus to efficient adaptation, as seen in EndoDAC \cite{endodac}, which uses parameter-efficient techniques like LoRA\cite{lora} to fine-tune large pre-trained models for surgical data. While effective, this maintained a non-unified structure. More recently, the pursuit of geometrically coherent online systems led to unified frameworks like Endo3R \cite{endo3r}. It extended pairwise reconstruction to long, dynamic videos using an uncertainty-aware memory mechanism and output a single, globally aligned pointmap from which both scale-consistent depth and camera poses are derived, ensuring inherent consistency. Our work specifically adapts a SOTA general-domain model, CUT3R~\cite{cut3r}, to the unique challenges of the surgical environment.
 
\section{Methodology}

\begin{figure*}[t]
    \centering
    \includegraphics[width=\textwidth]{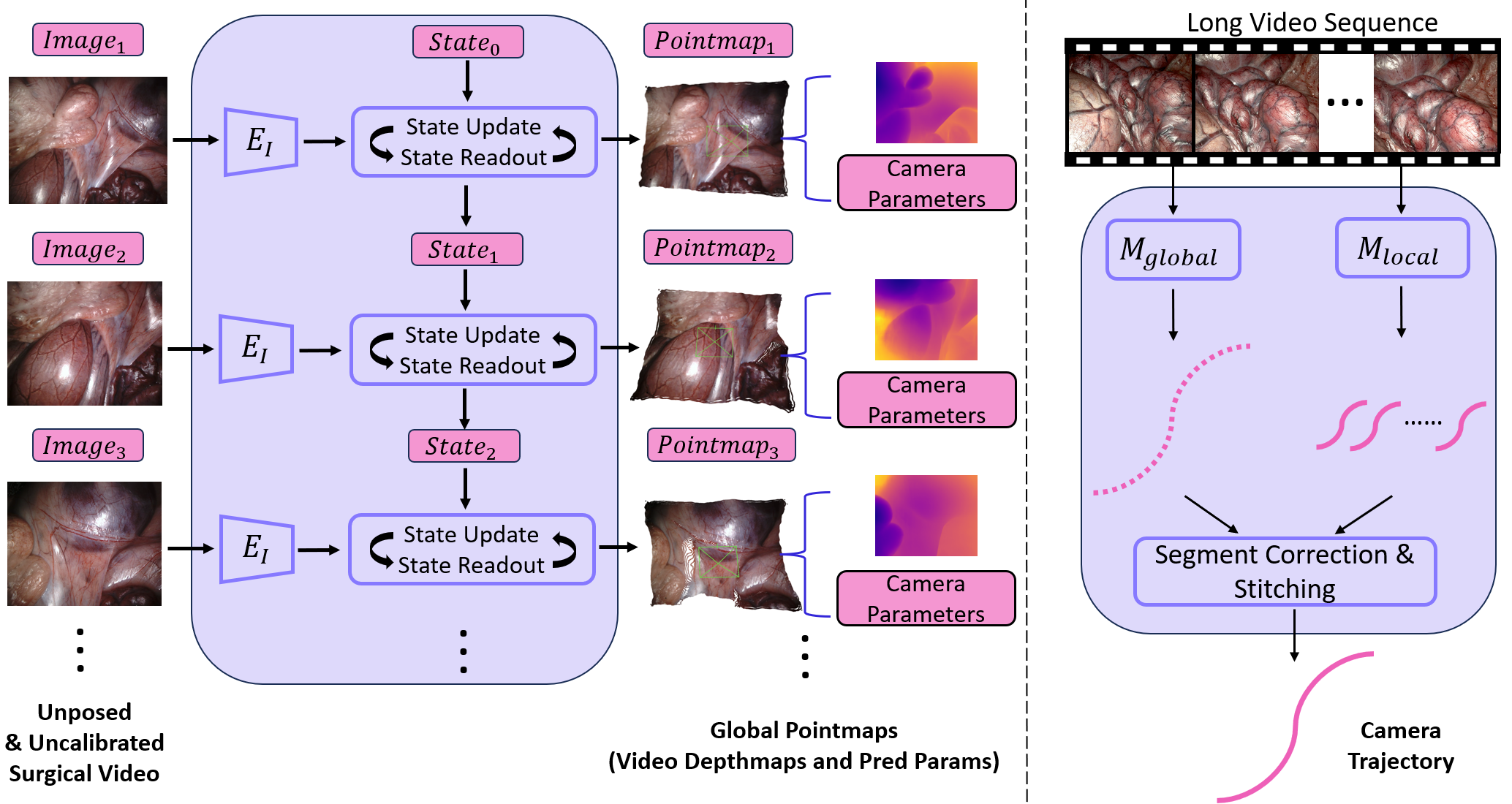}
    \vspace{1mm}
    \caption{
        \textbf{Overview of SurgCUT3R.} 
        \textbf{Left:} The unified reconstruction pipeline. Streaming video frames are encoded via a ViT encoder and interact with a persistent state, which is continuously updated to sequentially output the pointmap and camera parameter for each frame. 
        \textbf{Right:} Our hierarchical framework for long-sequence inference. The pink lines represent camera trajectories. A sparse but globally stable trajectory from a global model ($M_{global}$) provides anchor points to correct and stitch the dense but locally drifting trajectories from a local model ($M_{local}$), producing a final, drift-corrected trajectory.
    }
    \label{fig:pipeline}
\end{figure*}


Our proposed method, SurgCUT3R, establishes a systematic framework for adapting a unified reconstruction model to the challenging domain of monocular surgical video, with the main objective of achieving geometrically consistent results. To this end, we introduce a solution built on three key contributions: (1) To address the scarcity of supervised training data, we develop a pseudo-GT generation pipeline that leverages existing stereo datasets to produce high-quality depth labels. (2) To ensure robustness against inherent data imperfections, we employ a hybrid supervision strategy that couples our pseudo-GT data with a geometric self-correction mechanism. (3) To tackle the challenge of long-sequence inference, we propose a hierarchical framework that effectively suppresses pose drift. An overview of our method is illustrated in Fig. \ref{fig:pipeline}.

\subsection{Preliminaries: CUT3R\cite{cut3r}}

CUT3R \cite{cut3r} is a unified online reconstruction framework. Its core feature is the use of a continuously updated state to process an incoming stream of images. The model generates scale-consistent 3D pointmaps and camera parameters for each new image frame, progressively accumulating these pointmaps to build a coherent and dense 3D scene reconstruction.

Its core module is a State-Input interaction mechanism, where the model maintains a set of tokens as a latent state, denoted by $s$. Before processing any image, this state is initialized as a set of learnable tokens. For each input image $I_t$ at timestep $t$, a Vision Transformer encoder first converts it into image tokens $F_t$. 

These tokens then enter the main interaction process, which is visualized in the left panel of Fig. \ref{fig:pipeline}. Here, the tokens $F_t$ engage in a bidirectional interaction with the previous state, $s_{t-1}$, within the module labeled as `State Update State Readout'. This module performs two simultaneous operations: the `state-update' operation integrates new information from $I_t$ to produce the updated state $s_t$. Concurrently, the `state-readout' operation enriches the image tokens $F_t$ with historical context from $s_{t-1}$ to create enhanced tokens $F_t'$. While not explicitly shown in the figure, a learnable `pose token' ($z$) is processed alongside $F_t$, and its output ($z_t'$) is used to capture global scene information for predicting the camera pose. The original CUT3R \cite{cut3r} paper formulates this entire interaction as:

\begin{equation}
    [z_{t}^{\prime}, F_{t}^{\prime}], s_{t} = \text{Decoder}([z, F_t], s_{t-1})
\end{equation}

Following the interaction, the model extracts explicit 3D representations from the enhanced tokens using several heads. Specifically, the model predicts pointmaps and their corresponding confidence maps in two coordinate frames: $(\hat{X}_{t}^{\text{self}}, C_{t}^{\text{self}})$ in the input image's camera coordinate system, and $(\hat{X}_{t}^{\text{world}}, C_{t}^{\text{world}})$ in the world coordinate system (defined by the first frame's camera). Concurrently, the model predicts the camera pose $\hat{P}_{t}$, which represents the transformation from the current frame to the world frame:
\begin{align}
    (\hat{X}_{t}^{\text{self}}, C_{t}^{\text{self}}) &= \text{Head}_{\text{self}}(F_{t}^{\prime}) \\
    (\hat{X}_{t}^{\text{world}}, C_{t}^{\text{world}}) &= \text{Head}_{\text{world}}(F_{t}^{\prime}, z_{t}^{\prime}) \\
    \hat{P}_{t} &= \text{Head}_{\text{pose}}(z_{t}^{\prime})
\end{align}
where $\text{Head}_{\text{self}}$ and $\text{Head}_{\text{world}}$ adopt the DPT\cite{dpt} architecture, and $\text{Head}_{\text{pose}}$ is an MLP network. 

During the training phase, the model is optimized using a composite loss function. For pointmap regression, a confidence-aware regression loss is applied:
\begin{equation}
    \mathcal{L}_{\text{conf}} = \sum_{(\hat{x},c) \in (\hat{\mathcal{X}}, C)} \left( c \cdot \left\| \frac{\hat{x}}{\hat{s}} - \frac{x}{s} \right\|_{2} - \alpha \log c \right)
    \label{equ:Lconf}
\end{equation}
where $\hat{s}$ and $s$ are scale normalization factors for the predicted point set $\hat{\mathcal{X}}$ and GT $\mathcal{X}$ respectively. For the camera pose, the model minimizes the L2 norm between the prediction and the GT:
\begin{equation}
    \mathcal{L}_{\text{pose}} = \sum_{t=1}^{N} \left( \|\hat{q}_t - q_t\|_{2} + \left\|\frac{\hat{\tau}_t}{\hat{s}} - \frac{\tau_t}{s}\right\|_{2} \right)
    \label{equ:Lpose}
\end{equation}
where the pose $\hat{P}_t$ is parameterized by a quaternion $\hat{q}_t$ and a translation vector $\hat{\tau}_t$. 

\subsection{Proposed Method: SurgCUT3R}
\subsubsection{Pseudo-GT Generation for Surgical Scenes}
\label{sec:pseudo_gt}

A primary challenge in applying supervised 3D reconstruction models like CUT3R\cite{cut3r} to specialized medical field is the scarcity of suitable training data. These models require large-scale datasets with per-frame GT for both camera pose and dense depth. While existing surgical datasets such as SCARED \cite{scared} and StereoMIS \cite{stereomis} provide stereo video sequences with GT camera parameters, they critically lack dense and reliable depth information for every frame. The SCARED\cite{scared} dataset only contains sparse structured-light depth for the initial frame of each sequence, while StereoMIS\cite{stereomis} offers no GT depth at all. This data gap presents a fundamental barrier to fine-tuning SOTA unified reconstruction models for the surgical domain.

To overcome this obstacle, we propose a pseudo-GT generation pipeline designed to synthesize a high-fidelity dataset for supervised training. The core principle of our pipeline is to leverage the inherent geometric constraints of the available stereo video data to generate dense and metric-scale pseudo-GT depth maps for monocular training.

Our pipeline transforms the raw stereo sequences into a collection of (image, pseudo-GT depth, and GT pose) triplets through a systematic two-step process, as illustrated in Fig.~\ref{fig:gt_generation}:

\begin{figure}[t]
    \centering
    \includegraphics[width=1\columnwidth]{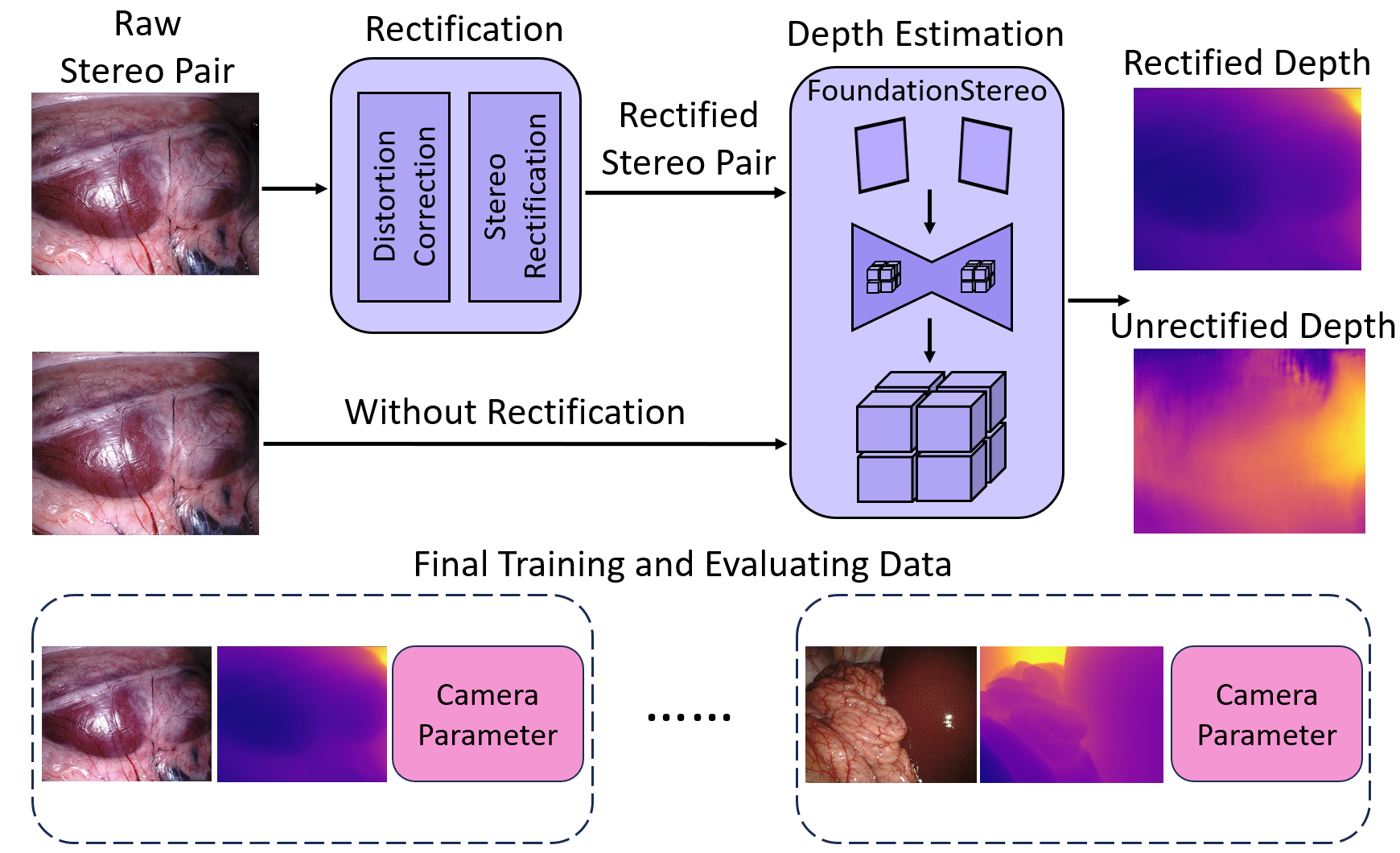}
    \vspace{1mm}
    \caption{\textbf{Pipeline of pseudo GT depth generation.} 
    The process rectifies the raw stereo pair before feeding it into FoundationStereo~\cite{foundationstereo} to generate a geometrically correct and metric-scale depth map. }
    \label{fig:gt_generation}
\end{figure}

\paragraph{Stereo Preprocessing and Rectification.}
We utilize the stereo video sequences from the SCARED~\cite{scared} and StereoMIS \cite{stereomis} datasets. Some of the raw endoscopic images in these datasets suffer from non-linear lens distortions and are not co-planar, resulting in skewed epipolar lines that complicate stereo matching. To address this, we follow the preprocessing procedures outlined in MSDESIS~\cite{msdesis} and Endo-4DGS \cite{endo4dgs}. The process, guided by the provided camera calibration files, involves two main steps: first, we perform distortion correction to remove lens-induced artifacts. Second, we apply stereo rectification to align the image planes. This procedure yields the final distortion-free and row-aligned stereo image pairs, $I_L$ and $I_R$, required for reliable stereo matching.

\paragraph{Metric-Scale Depth Synthesis and Dataset Assembly.}
We generate dense depth maps from the rectified stereo pairs, $I_L$ and $I_R$, using FoundationStereo~\cite{foundationstereo}. The resulting disparity map , denoted as $d$, is converted to a metric-scale depth map $D$ using the known camera baseline $b$ and focal length $f$ which are provided by the dataset, according to the equation \ref{eq:depth_from_disparity}:
\begin{equation}
    D(u, v) = \frac{b \cdot f}{d(u, v)}.
    \label{eq:depth_from_disparity}
\end{equation}
Finally, for each frame at time $t$, we assemble the left image $I_L^{(t)}$, the synthesized depth map $D^{(t)}$, and the GT camera pose and intrinsics $P_{\text{GT}}^{(t)}$ to form the training dataset. This pipeline produces a large metric-scale dataset suitable for supervised training.

\definecolor{Gray}{gray}{0.9}

\begin{table*}[t]
  \centering
  \setlength{\tabcolsep}{4pt} 
  \caption{QUANTITATIVE EVALUATION OF OUR SURGCUT3R METHOD AGAINST EXISTING METHODS IN ENDOSCOPIC SCENE RECONSTRUCTION. THE OPTIMAL AND SUBOPTIMAL RESULTS ARE SHOWN IN BOLD AND UNDERLINED RESPECTIVELY. }
  \label{tab:performance_comparison}
  \begin{tabular}{@{}c|c|c|ccccc|cc|c@{}}
    \toprule
    \textbf{Dataset} & \textbf{Category} & \textbf{Method} & \textbf{Abs Rel}$\downarrow$ & \textbf{Sq Rel}$\downarrow$ & \textbf{RMSE}$\downarrow$ & \textbf{RMSE LOG}$\downarrow$ & $\boldsymbol{\delta < 1.25}\uparrow$ & \textbf{ATE}$\downarrow$ & \textbf{RTE}$\downarrow$ & \textbf{FPS}$\uparrow$ \\
    \midrule
    
    \multirow{6}{*}{SCARED\cite{scared}} 
    & \multirow{2}{*}{Optimization-based} & MonST3R(w/ Opt)\cite{monst3r}    & \underline{0.098} & \underline{1.237} & \underline{7.979}  & \underline{0.132} & \underline{0.904} & \underline{21.774} & \underline{1.582} & \underline{0.3}  \\
    &                            & MegaSaM\cite{megasam}    & \textbf{0.056} & \textbf{0.392} & \textbf{4.586}  & \textbf{0.074} & \textbf{0.978} & \textbf{2.002}  & \textbf{0.315} & \textbf{0.7}  \\
    \cmidrule(l){2-11} 
    & \multirow{4}{*}{Feed-forward} & Spann3R\cite{spann3r}    & 0.119 & 2.524 & 10.218 & 0.148 & 0.867 & {10.258} & 1.260 & 19.2 \\
    &                            & AF-SfMLearner\cite{afsfm} & 0.073 & 0.534 & 5.028  & 0.081 & 0.964 & 10.312 & 0.971 & 3.6  \\
    &                            & EndoDAC\cite{endodac}    & \underline{0.059} & \underline{0.443} & \underline{4.833}  & \underline{0.079} & \underline{0.973} & \underline{10.225} & \underline{0.963} & \textbf{36.3} \\
    &                            & \cellcolor{Gray}\textbf{SurgCUT3R (Ours)} & \cellcolor{Gray}\textbf{0.057} & \cellcolor{Gray}\textbf{0.410} & \cellcolor{Gray}\textbf{4.647} & \cellcolor{Gray}\textbf{0.077} & \cellcolor{Gray}\textbf{0.977} & \cellcolor{Gray}\textbf{5.514} & \cellcolor{Gray}\textbf{0.752} & \cellcolor{Gray}\underline{19.7} \\
    
    \midrule
    
    \multirow{6}{*}{StereoMIS\cite{stereomis}} 
    & \multirow{2}{*}{Optimization-based} & MonST3R(w/ Opt)\cite{monst3r}    & \underline{0.187} & \underline{4.342} & \underline{15.671} & \underline{0.225} & \underline{0.672} & \underline{41.183} & \underline{1.412} & \underline{0.3}  \\
    &                            & MegaSaM\cite{megasam}    & \textbf{0.061} & \textbf{0.506} & \textbf{4.615}  & \textbf{0.084} & \textbf{0.976} & \textbf{19.705} & \textbf{0.877} & \textbf{0.7}  \\
    \cmidrule(l){2-11}
    & \multirow{4}{*}{Feed-forward} & Spann3R\cite{spann3r}    & 0.213 & 7.075 & 17.535 & 0.318 & 0.693 & 29.842 & 1.391 & 19.2 \\
    &                            & AF-SfMLearner\cite{afsfm} & 0.102 & 0.872 & 7.410  & 0.110 & 0.896 & \underline{25.128} & 1.197 & 3.6 \\
    &                            & EndoDAC\cite{endodac}    & \underline{0.075} & \underline{0.672} & \underline{6.047}  & \underline{0.094} & \underline{0.957} & \textbf{24.264} & \underline{1.121} & \textbf{36.3} \\
    &                            & \cellcolor{Gray}\textbf{SurgCUT3R (Ours)} & \cellcolor{Gray}\textbf{0.070} & \cellcolor{Gray}\textbf{0.637} & \cellcolor{Gray}\textbf{5.732} & \cellcolor{Gray}\textbf{0.091} & \cellcolor{Gray}\textbf{0.965} & \cellcolor{Gray}25.939 & \cellcolor{Gray}\textbf{0.902} & \cellcolor{Gray}\underline{19.7} \\
    
    \bottomrule
  \end{tabular}
\end{table*}

\subsubsection{Hybrid Supervision for Robust Training}
\label{sec:hybrid_supervision}

While the pseudo-GT pipeline described in Section \ref{sec:pseudo_gt} provides high-quality metric-scale supervision, the synthesized depth maps are not entirely free from imperfections. Surgical scenes frequently contain challenging elements such as specular reflections from wet tissue surfaces, smoke from electrocautery, and low-texture regions. These factors can introduce local noise and inaccuracy into the depth maps generated by the stereo matching model. Directly training a network with a purely supervised loss on this imperfect data risks overfitting to label noise, which could lead the model to learn incorrect geometric priors.

To mitigate the effect of label noise and enhance the model's geometric understanding, we introduce a hybrid supervision strategy for training. This strategy combines the direct supervision from our pseudo-GT data with a geometric consistency self-supervision term. The core idea is to use the pseudo-GT to guide the overall learning while leveraging multi-view self-consistency as a powerful regularizer to refine the geometric structure and promote robustness against noisy labels.

Our total training objective, $\mathcal{L}_{\text{total}}$, is a weighted sum of the supervised losses from the baseline model and our self-supervised consistency loss:
\begin{equation}
    \mathcal{L}_{\text{total}} = (\mathcal{L}_{\text{conf}} + \mathcal{L}_{\text{pose}}) + \lambda_{\text{consist}} \cdot \mathcal{L}_{\text{consistency}},
    \label{eq:total_loss}
\end{equation}

where $\lambda_{\text{consist}}$ is a hyperparameter that balances the supervised and self-supervised terms.

\paragraph{Supervised Terms ($\mathcal{L}_{\text{conf}}$ and $\mathcal{L}_{\text{pose}}$).}
The primary component of our loss is the supervised objective inherited from the baseline CUT3R\cite{cut3r} model, comprising two parts. The first is a confidence-weighted regression loss, $\mathcal{L}_{\text{conf}}$ (Eq. \ref{equ:Lconf}), which minimizes the error between the predicted and pseudo-GT point clouds. The second is a pose regression loss, $\mathcal{L}_{\text{pose}}$ (Eq. \ref{equ:Lpose}), which minimizes the error between the predicted and GT camera pose. These terms anchor the model's predictions to the largely accurate and metric-scale pseudo-GT we have generated, ensuring the model learns the overall correct scale and structure of the surgical scene.

\paragraph{Self-Supervised Term ($\mathcal{L}_{\text{consistency}}$).}
To mitigate the effect of label noise, we introduce a comprehensive self-supervised consistency loss $\mathcal{L}_{\text{consistency}}$, inspired by the depth optimization objective in MegaSaM \cite{megasam}. While MegaSaM\cite{megasam} employs this loss for post-process refinement, we adapt it as a self-supervised signal during training. The necessary inputs for this loss, such as per-frame depth maps and relative poses, are taken directly from our model's own predictions within the training batch. The loss is a composite of three main terms:
\begin{equation}
    \mathcal{L}_{\text{consistency}} = w_{\text{flow}}\mathcal{C}_{\text{flow}} + w_{\text{temp}}\mathcal{C}_{\text{temp}} + w_{\text{prior}}\mathcal{C}_{\text{prior}},
    \label{eq:consistency_loss_components}
\end{equation}
where $w$ are weighting coefficients. The components are defined as follows:

\begin{itemize}
    \item \textbf{Optical Flow Consistency ($\mathcal{C}_{\text{flow}}$):} This term enforces consistency between the 2D optical flow $\text{flow}_{i \to j}$, which is computed by a pre-trained RAFT model \cite{raft} and the 2D motion field induced by the model's predicted depth $\hat{D}_i$ and relative pose $\hat{G}_{ij}$. For a pixel $\mathbf{p}$ in frame $i$, its corresponding point in frame $j$ is $\mathbf{p}' = \mathbf{p} + \text{flow}_{i \to j}(\mathbf{p})$. The reprojection of $\mathbf{p}$ using the model's prediction is $u_{ij}(\mathbf{p})$. The loss is explained as follows:
    \begin{equation}
        \mathcal{C}_{\text{flow}} = \| u_{ij}(\mathbf{p}) - \mathbf{p}' \|_1.
        \label{equ:Cflow}
    \end{equation}
    
    \item \textbf{Temporal Geometric Consistency ($\mathcal{C}_{\text{temp}}$):} This loss encourages the 3D structure to be consistent over time. It measures the scale-invariant error between a point's depth $\hat{D}_j(\mathbf{p'})$ in the target frame $j$ at the flow-warped coordinate $\mathbf{p}'$, and its depth $P_z^{i \to j}$ projected from the source frame $i$. The loss is explained as follows:
    \begin{equation}
        \mathcal{C}_{\text{temp}} = \left\| \max\left(\frac{P_z^{i \to j}(\mathbf{p})}{\hat{D}_j(\mathbf{p'})}, \frac{\hat{D}_j(\mathbf{p'})}{P_z^{i \to j}(\mathbf{p})}\right) - 1 \right\|_1.
        \label{equ:Ctemp}
    \end{equation}
    
    \item \textbf{Prior Regularization ($\mathcal{C}_{\text{prior}}$):} This term regularizes the geometry to prevent drift and maintain surface properties. It is a composite of three sub-terms from MegaSaM\cite{megasam}: a scale-invariant loss $\mathcal{C}_{\text{si}}$ to maintain overall shape, a multi-scale gradient matching loss $\mathcal{C}_{\text{grad}}$ to preserve geometric details, and a surface normal consistency loss $\mathcal{C}_{\text{normal}}$ to encourage locally smooth surfaces.
\end{itemize}
It is worth noting that the original MegaSaM\cite{megasam} utilizes a predicted uncertainty map to apply per-pixel weights within the photometric and temporal consistency losses, which down-weights contributions from dynamic regions. Since our training is conducted on static surgical scenes, we do not predict this map and instead use a constant value for this uncertainty weights.

The supervised loss provides the strong and scale-consistent signal necessary for accuracy, while the self-supervision loss acts as a geometric regularizer, enabling the model to self-correct from imperfections in the pseudo-GT data. This combination enhances the model's robustness, leading to a more accurate and geometrically coherent final reconstruction.

\subsubsection{Hierarchical Framework for Long-Sequence Inference}
\label{sec:hierarchical_framework}

A significant challenge for autoregressive reconstruction models like CUT3R\cite{cut3r} is their performance degradation over long video sequences. As the model processes frames sequentially, small prediction errors in pose inevitably accumulate. This accumulation of errors manifests itself as pose drift, where the estimated camera trajectory gradually deviates from the true path. This issue severely limits the applicability of such models for tracking entire surgical procedures, which are often lengthy.

To address the challenge of long-sequence tracking, we propose a hierarchical framework for pose estimation. The core idea is to establish a globally consistent camera trajectory by correcting the drift of a dense, short-term tracker using a sparse, long-term tracker as a stable reference. Our method achieves this through per-segment alignment and an error correction module, effectively suppressing error accumulation.

This hierarchical framework features a global model, a local model, and a fusion pipeline that integrates their outputs to generate a drift-corrected camera trajectory.


\begin{figure*}[t]
    \centering

    \includegraphics[width=\textwidth]{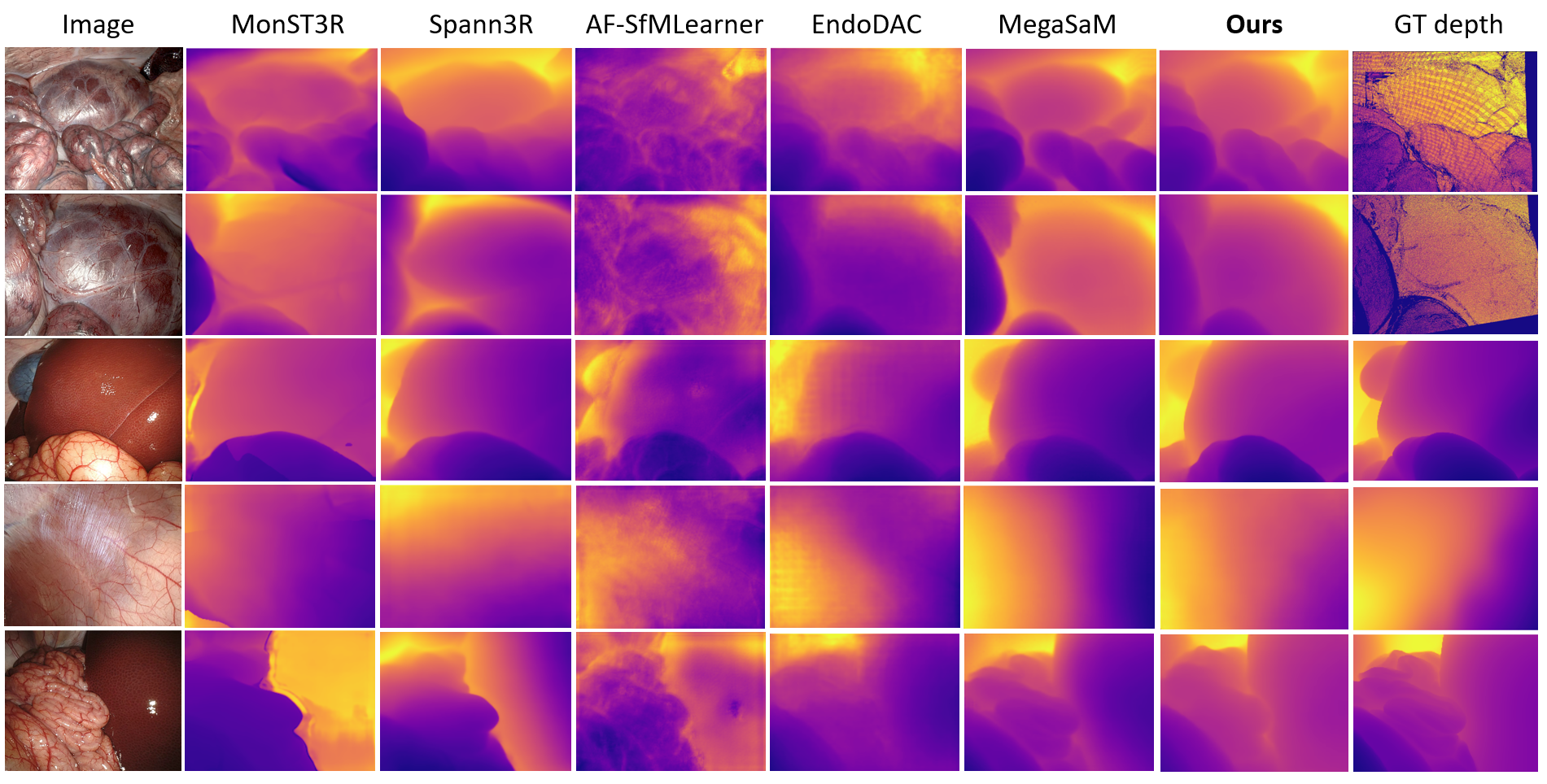}

    \vspace{2mm} 
    
    \caption{\textbf{Qualitative results of monocular depth estimation.} We compare our method with MonST3R\cite{monst3r}, Spann3R\cite{spann3r}, AF-SfMLearner\cite{afsfm}, EndoDAC\cite{endodac} and MegaSaM\cite{megasam} on SCARED\cite{scared} and StereoMIS\cite{stereomis} datasets. Our method achieves the best qualitative results in feed-forward methods.}
    \label{fig:qualitative_results}
\end{figure*}


\paragraph{Dual-Model Specialization.}
We train two instances of our model with different temporal sampling strategies to specialize them for different tasks:
\begin{itemize}
    \item \textbf{Global Model ($\mathcal{M}_{\text{global}}$):} This model is trained on sparsely sampled video frames (e.g., with a max interval of 12 frames). Its purpose is to learn robust long-range motion estimation, focusing on global trajectory consistency.
    \item \textbf{Local Model ($\mathcal{M}_{\text{local}}$):} This model is trained on densely sampled video frames (e.g., with a max interval of 3 frames). It is specialized for capturing accurate relative camera motion within a short time window.
\end{itemize}

\paragraph{Trajectory Correction and Stitching.}
Our pipeline corrects the trajectory on a per-segment basis as follows:
\begin{itemize}
    \item {\textbf{Pose Generation:}} We first generate two sets of poses: a sparse but globally stable "anchor" trajectory using $\mathcal{M}_{\text{global}}$, and multiple dense but locally drifting pose sequences for each segment between anchors using $\mathcal{M}_{\text{local}}$.
    \item {\textbf{Per-Segment Correction:}} For each local segment, we first align the segment's starting pose to its corresponding global anchor. Then, we calculate the drift error, which is the discrepancy between the segment's aligned end pose and the next global anchor. This error is then distributed across all frames within the segment by interpolating the rotational and translational components using spherical linear interpolation and linear interpolation, respectively. This process ultimately yields the complete drift-corrected trajectory for the entire long sequence.
\end{itemize}

This hierarchical framework effectively mitigates pose drift in long-sequence tracking. By combining the long-term stability of a global model with the local accuracy of a dense model, our method generates a complete drift-corrected trajectory.

\section{Experiments}
\label{sec:experiments}

\subsection{Experiment Setting}
\label{sec:exp_setting}

\subsubsection{Datasets}
We train and evaluate our method on two public surgical datasets: SCARED \cite{scared} and StereoMIS \cite{stereomis}. For {training}, we use the SCARED\cite{scared} dataset, following the official split in SCARED\cite{scared} challenge (i.e., Datasets 1--7 for training and Datasets 8--9 for testing). However, a prior study~\cite{revisiting} noted significant calibration errors in Dataset4 and Dataset5. To ensure that our model learns from high-quality data, we exclude these two subsets from our training dataset. To evaluate the generalization ability, we test on 4 unseen StereoMIS\cite{stereomis} sequences for cross-dataset validation: Sequence 1 (frames 6900--7600 in P2-1), Sequence 2 (frames 1200--1600 in P2-4), Sequence 3 (frames 9300--10200 in P2-5), and Sequence 4 (frames 1000--1300 in P2-8). These sequences were selected from our generated pseudo-GT data for their significant scene dynamics and camera motion, as well as for their high reconstruction quality with minimal visual artifacts like smoke or glare.

\subsubsection{Evaluation Metrics}
We evaluate performance across three aspects: pose accuracy, depth estimation quality, and efficiency. For {pose accuracy}, we use Absolute Trajectory Error (ATE) for global consistency and Relative Trajectory Error (RTE) with a window size of 16 for local accuracy. The unit for both ATE and RTE is millimeter (mm). For {depth quality}, we follow EndoDAC \cite{endodac} and report five standard metrics: Absolute Relative Error (Abs Rel), Square Relative Error (Sq Rel), Root Mean Square Error (RMSE), RMSE log, and the accuracy metric $\delta < 1.25$. The evaluation is performed at a resolution of 256x192. GT depth for SCARED\cite{scared} is from the dataset itself, while for StereoMIS\cite{stereomis}, we use our generated pseudo-GT depth. For {efficiency}, we report the inference speed in Frames Per Second (FPS).

\subsubsection{Implementation Details}
The experiments were conducted on an NVIDIA GPU 4090 with PyTorch framework. We use the AdamW optimizer with a learning rate of $1.0 \times 10^{-5}$, a weight decay of 0.05, and a batch size of 8. Our training follows a two-stage strategy. First, we fine-tune the publicly available CUT3R\cite{cut3r} weight for 5 epochs using only the supervised losses ($\mathcal{L}_{\text{conf}} + \mathcal{L}_{\text{pose}}$). Then, we introduce the self-supervised term $\mathcal{L}_{\text{consistency}}$ and continue training for two more epochs to refine the model. For the hierarchical inference framework, both the global and local models share the same training configuration, but are differentiated by their maximum temporal sampling interval during training, which is set to 12 for the global model and 3 for the local model.

\subsection{Qualitative and Quantitative Evaluation}
\label{sec:exp_results}

We evaluate our method by comparing its performance with several SOTA methods on both the SCARED\cite{scared} and StereoMIS\cite{stereomis} datasets.

\paragraph{Quantitative Comparison.}
Table \ref{tab:performance_comparison} highlights our method's optimal balance between accuracy and efficiency. On the SCARED\cite{scared} dataset, while MegaSaM\cite{megasam} achieves the highest depth and pose accuracy, its heavy computational cost limits it to 0.7 FPS. In contrast, our method delivers near-SOTA depth and second-best pose estimation at a significantly faster 19.7 FPS. This competitive performance generalizes well to the unseen StereoMIS\cite{stereomis} dataset, establishing our approach as a highly practical, near-real-time solution for surgical reconstruction. Since relying on self-generated pseudo-GT for StereoMIS\cite{stereomis} evaluation introduces potential bias, future validation on dynamic datasets with authentic ground truth is needed.

\paragraph{Qualitative Comparison.}
Fig. \ref{fig:qualitative_results} visually validates our depth estimation, showing that SurgCUT3R achieves precise depth maps with superior relative scale. And as shown in Fig. \ref{fig:reconstruction_examples}, The high quality of the resulting 3D reconstructions and pose estimations is mainly attributed to our accurate depth and pose predictions.

\subsection{Ablations and Analysis}
\label{sec:ablations}

To validate the effectiveness of our proposed components, we conduct two ablation studies on the SCARED\cite{scared} dataset.

\paragraph{Effect of Hybrid Supervision.}
We first evaluate the impact of our self-supervised consistency loss. As shown in Table \ref{tab:ablation_loss}, our consistency loss (w/ $L_{\text{consistency}}$) yields a marginal but consistent improvement across depth estimation metrics compared to using only supervised losses (w/o $L_{\text{consistency}}$). 

\begin{table}[h]
  \centering
  \caption{ABLATION STUDY ON THE LOSS FUNCTION}
  \label{tab:ablation_loss}
  {
  \setlength{\tabcolsep}{10pt} 
  \begin{tabular}{@{}lccc@{}}
    \toprule
    \textbf{Configuration} & \textbf{Sq Rel}$\downarrow$ & \textbf{RMSE}$\downarrow$ & $\boldsymbol{\delta < 1.25}\uparrow$ \\
    \midrule
    w/o $L_{\text{consistency}}$ & 0.423 & 4.763 & 0.975 \\
    \textbf{w/ $L_{\text{consistency}}$ (Ours)} & \textbf{0.410} & \textbf{4.647} & \textbf{0.977} \\
    \bottomrule
  \end{tabular}
  }
\end{table}

\begin{figure}[h]
    \centering
    \includegraphics[width=\columnwidth]{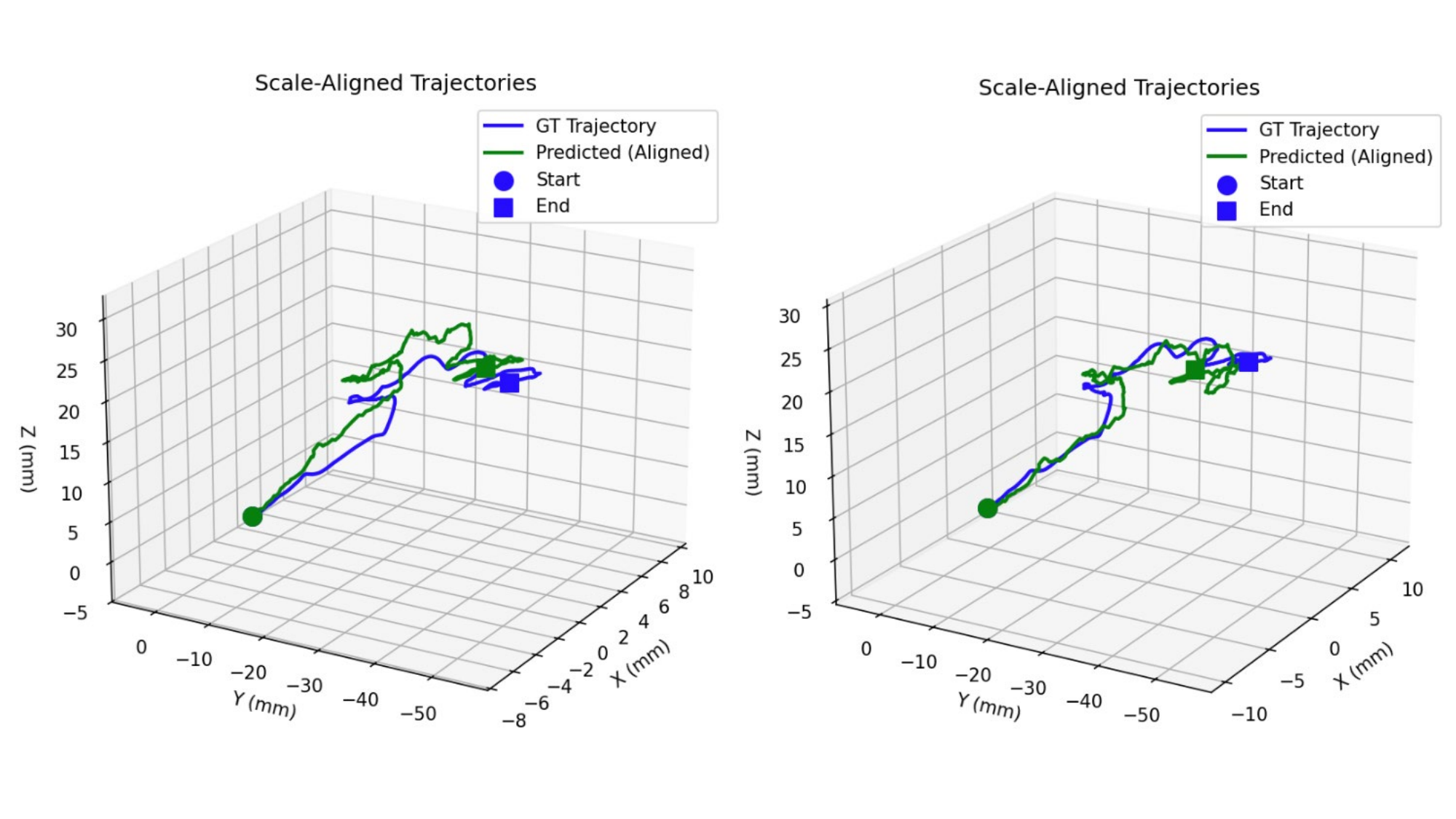}
    \caption{Qualitative comparison of camera trajectories. \textbf{Left:} Without the hierarchical inference framework. \textbf{Right:} With our hierarchical inference framework (Ours).}
    \label{fig:ablation_trajectory}
\end{figure}

\paragraph{Effect of Hierarchical Framework.}
Next, we analyze the contribution of our dual-model hierarchical framework for long-sequence pose estimation. Table \ref{tab:ablation_architecture} shows that using our dual-architecture significantly reduces the ATE compared to using a single model. This result quantitatively demonstrates the effectiveness of our approach in mitigating accumulated pose drift. The qualitative improvement in trajectory stability is further visualized in Fig. \ref{fig:ablation_trajectory}.

\begin{table}[h]
  \centering
  \caption{ABLATION STUDY ON THE ARCHITECTURE}
  \label{tab:ablation_architecture}
  { 
  \setlength{\tabcolsep}{26pt} 
  \begin{tabular}{lc} 
    \toprule
    \textbf{Configuration} & \textbf{ATE}$\downarrow$ \\
    \midrule
    CUT3R\cite{cut3r} Only      & 9.361 \\
    \textbf{Dual-Arch (Ours)} & \textbf{5.514} \\
    \bottomrule
  \end{tabular}
  } 
\end{table}

\section{Conclusion}
\label{sec:conclusion}

We present SurgCUT3R, a unified framework for monocular surgical 3D reconstruction. To overcome data scarcity and inherent label noise, we develop a metric-scale pseudo-GT generation pipeline from stereo datasets, coupled with a hybrid supervision strategy for geometric self-correction. Furthermore, a hierarchical inference framework is introduced to effectively mitigate long-term pose drift. Unlike slow offline methods such as MegaSaM\cite{megasam}, SurgCUT3R strikes a clinically practical balance, delivering competitive accuracy at 19.7 FPS. This efficiency makes it a robust solution for surgical navigation.
Although our current pseudo-GT is effective, future work will take advantage of offline optimization frameworks such as MegaSaM~\cite{megasam} to mitigate artifact-induced depth misalignments and construct more accurate training data.






\clearpage
\bibliographystyle{class/IEEEtran}
\bibliography{class/IEEEabrv,class/reference}
   
\end{document}